\documentclass[conference]{IEEEtran}
\IEEEoverridecommandlockouts
\usepackage{cite}
\usepackage{amsmath,amssymb,amsfonts}
\usepackage{algorithmic}
\usepackage{graphicx}
\usepackage{subcaption}
\usepackage{textcomp}
\usepackage{xcolor}
\usepackage{siunitx}
\usepackage{soul}
\usepackage{hyperref}
\usepackage[T1]{fontenc}
\def\BibTeX{{\rm B\kern-.05em{\sc i\kern-.025em b}\kern-.08em
    T\kern-.1667em\lower.7ex\hbox{E}\kern-.125emX}}
\begin{document}

\title{Adversarial Reinforcement Learning for Procedural Content Generation}

\author{\IEEEauthorblockN{}
\IEEEauthorblockA{Linus Gisslén$^\dagger$, Andy Eakins$^\ddag$, Camilo Gordillo$^\dagger$, Joakim Bergdahl$^\dagger$, Konrad Tollmar$^\dagger$\\
\textit{SEED$^\dagger$, Frostbite$^\ddag$ - Electronic Arts (EA)}\\
lgisslen, aeakins, cgordillo, jbergdahl, ktollmar@ea.com}
}

\maketitle

\begin{abstract}
We present a new approach ARLPCG: Adversarial Reinforcement Learning for Procedural Content Generation, which procedurally generates and tests previously unseen environments with an auxiliary input as a control variable. Training RL agents over novel environments is a notoriously difficult task. One popular approach is to procedurally generate different environments to increase the generalizability of the trained agents. ARLPCG instead deploys an adversarial model with one PCG RL agent (called Generator) and one solving RL agent (called Solver). The Generator receives a reward signal based on the Solver's performance, which encourages the environment design to be challenging but not impossible. To further drive diversity and control of the environment generation, we propose using auxiliary inputs for the Generator. The benefit is two-fold: Firstly, the Solver achieves better generalization through the Generator's generated challenges. Secondly, the trained Generator can be used as a creator of novel environments that, together with the Solver, can be shown to be solvable. We create two types of 3D environments to validate our model, representing two popular game genres: a third-person platformer and a racing game. In these cases, we shows that ARLPCG has a significantly better solve ratio, and that the auxiliary inputs renders the levels creation controllable to a certain degree. For a video compilation of the results please visit \url{https://youtu.be/z7q2PtVsT0I}.
\end{abstract}


\section{Introduction}



Recent research in RL has made large strides in improvement with the introduction of deep RL. Many previously unsolved problems have been solved with this method and super-human performance has been achieved on Atari games, ViZDoom, StarCraft, Dota2, etc. \cite{berner2019dota, shao2019survey}. However, often the training and validation sets are practically identical in this kind of settings encouraging the agent to memorize the situation leading to poor generalization ability. For many areas of reinforcement learning, one challenge is to train adaptive agents that can handle previously unseen environments and situations. Areas including self-driving cars, computer games, robotics, etc. where the agents may encounter new situations frequently. 

Training an RL agent to solve a (fixed) game can indeed lead to super-human performance but with the risk of overfitting policies to the existing game which makes RL less effective on unseen environments \cite{packer2018assessing}. This is problematic from a game development perspective for several reasons. During development, the game, the environments, the characters, the assets, etc. can change on a day-to-day basis. One solution is to re-train the agents at every update but that is a costly, and often infeasible solution as the training could take longer than the iteration time of a development cycle. Furthermore, in game creation there are several reasons to have agents with high generalization ability. An agent that plays player created content (e.g. {\it Minecraft}, {\it SimCity}, and {\it The Sims}) requires a lot of adaptation and generalization to be able to solve a large plethora of challenges. Automated and/or assisted creation of assets in games is becoming an increasingly crucial component of game development as games become bigger and more complex. Most human created (manually, or assisted), as well as fully automated created assets, require testing. Ideally this should be done in real time to avoid inducing unnecessary delays in production. Even more crucially, there is a need for in-game AI to adapt to novel environments and situations, as it often interacts with human players or their creations, leading to an abundance of situations which it has not encountered in training. Therefore, the generalization ability of an agent is key to make RL a useful tool in games and game development.

In this paper, we will approach this problem by training agents on ever-changing environments in order to become more robust to unseen scenarios. As previously touched upon there are several use cases for this ability in games and game production:
\begin{itemize}
    \item In game development where agents need to be able to play previously unseen maps or characters, especially for automated game testing. 
    \item In games where agents need to be able to solve player generated maps. Here often there is no access to the training tools as they are not shipped in the game, thus re-training of the agents is impossible. 
    \item Real-time procedural content generation (PCG) where the environment may change according to various factors, e.g. player skill. 
\end{itemize}

It has been shown that PCG has a positive effect on immersiveness in games \cite{connor2017evaluating}. PCG techniques often focused on visuals and the generation of new textures, materials and assets such as trees, landscapes, etc. In this paper, we focus on the "playability" of assets, meaning that we want to create assets and environments that affect the way the game is played, not only perceived. This paper focuses on game development but we imagine this approach being useful in other areas where good generalization is needed.

\section{Previous Work}

Machine learning has been shown to potentially be a tool in automated game testing. Supervised learning (SL) can for example be used for rendered image glitch detection to classify bugs in the image output from a game \cite{ling2020using}. Recent studies on modern 3D games have shown the potential of RL to be used as game-AI \cite{harmer18}. Specifically, RL can be very beneficial in game testing as one feature of the RL agent is its ability to learn to play the game. As the game is learned from scratch without any prior knowledge on how to play or solve the game where even the controller has to be learned, no preconceived way of solving the task is chosen. This feature allows it to potentially identify unintended game behaviours and situations such as imbalances or exploits in the environment. Previous studies exploring the use of RL in game testing include combining evolutionary algorithms, RL and multi-objective optimization \cite{zheng2019wuji} and reinforcement learning for exploration and bug exploitation \cite{bergdahl2020augmenting}. 

PCG via ML refers to the use of machine learning to train models on existing game content, and then leverage these models to create novel content automatically. This assumes that there exists previous game content to train on. It has been found that in many cases, trained agents
overfit to the environment they were trained on \cite{risi2020increasing}. Therefore, PCG can be leveraged to increase generality of the trained agent as it has to learn to play a previously unseen environment \cite{risi2020increasing}. Furthermore, RL can be used for PCG by training level designing agents \cite{khalifa2020pcgrl}. 

An approach called generative playing networks (GPNs) \cite{bontrager2020fully} for generation and solving consists of three models: an environment generator G, an environment agent policy, and an environment value estimator. As each model is a differentiable function, GPNs are fully differentiable and generate content that is paired with a solver. Thus, this model gives the agent/solver control over the environment that is created. Similarly, the paired open-ended trailblazer (POET) algorithm is an environment-agent pair where the agent is trained on a parameterized environment. Here, open-ended learning was achieved by storing a population of environments and searching for new ones that are of appropriate difficulty for the agents. This generates increasingly complex and diverse learning environments paired with their solutions \cite{wang2019paired}.
The continuous teacher-student (CTS) employs a learning progress-based teacher algorithm where the teacher interacts with the student by generating a parameterized controlled environment. Here, a curriculum was created by explicitly modelling the difficulty of the goal space \cite{portelas2020teacher}.

There are several approaches and use cases to adversarial RL (ARL). Commonly, an adversarial agent is the opponent which plays against the training model, see e.g. \cite{uther1997adversarial}. Another model is to use ARL to make the agents more robust (robust adversarial RL (RARL)) \cite{pinto2017robust}, which trains an agent to operate in the presence of a destabilizing adversary RL agent that applies optimal disturbance forces to the system to maximize learning. 

\section{Model}
\label{sec:model}

\begin{figure}[!h]
    \centering
    \includegraphics[width=0.42\textwidth]{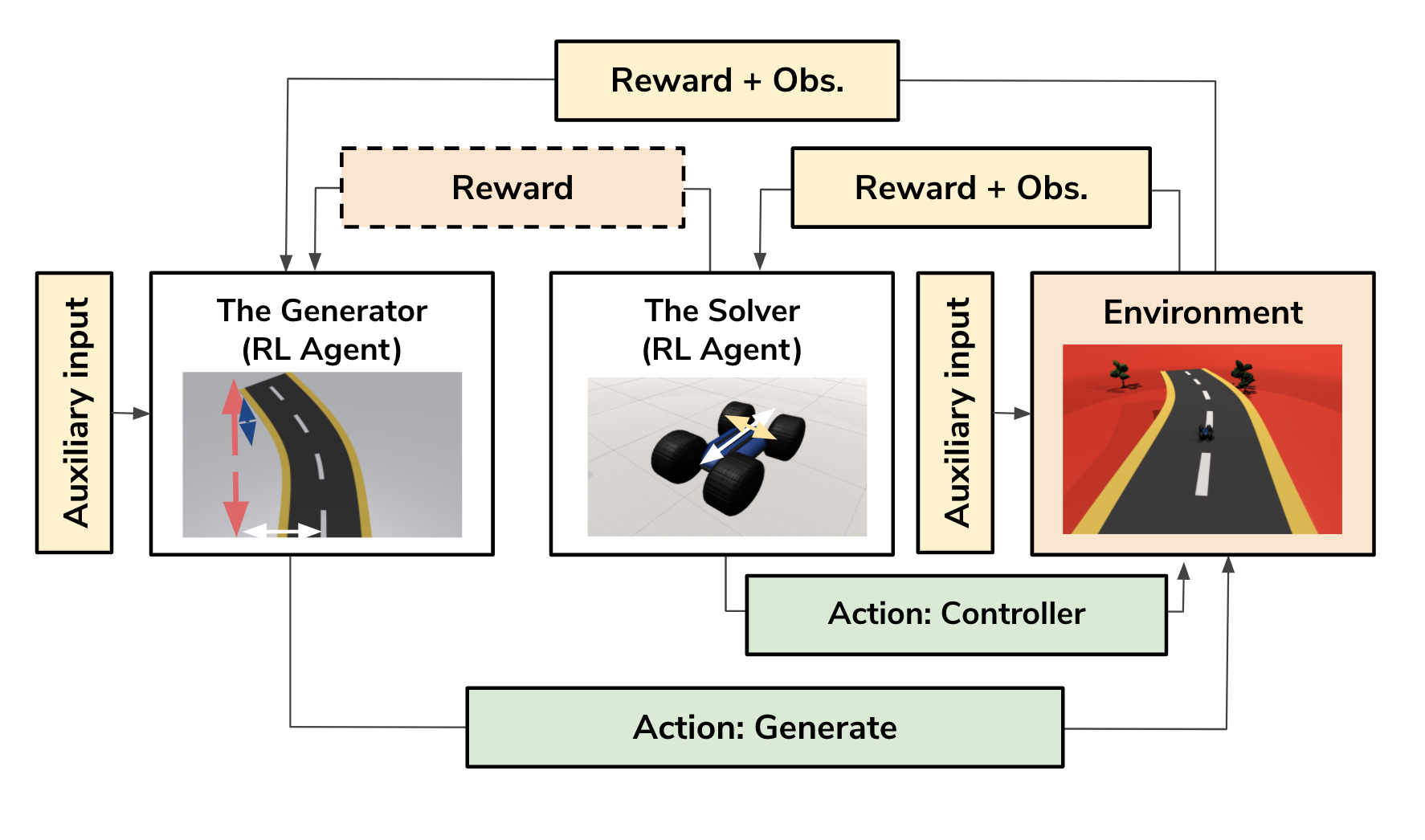}
    \caption{Schematic view of the architecture of adversarial RL PCG (ARLPCG). The model consists of two parts: One RL Generator and one RL Solver. The Generator receives a reward depending on the performance of the Solver as well as from the environment. The implicit reward (dashed) passed from Solver to Generator is merely conceptual as in practice all the rewards are passed from the environment. The auxiliary input is passed to both Generator (as an observation) and the Environment in order to let the environment adapt the reward function accordingly.}
    \label{fig:ARLPCGModel}
\end{figure}

The ARLPCG model draws inspiration from several ideas, below we list a few of them and their relation to our approach:
\begin{itemize}
    \item Training on procedurally generated content improves generalization in RL agents \cite{risi2020increasing}.
    \item RL can be used for PCG \cite{khalifa2020pcgrl}. This, combined with the first point, is a powerful idea which is fundamental to our model when it comes to creating ever changing environments for generalizing RL.
    \item Our Generator and Solver models can also be found in \cite{bontrager2020fully} where a semi-supervised model is composed of three parts: one Generator, one Solver, and one Evaluator (utility estimator). Their model requires that it is (practically) possible to encode simulator states into environment descriptions.
    \item Auxiliary input connected to a reward can be used as a training enhancer. Previous work shows that auxiliary inputs can be used in a pseudo-reward optimization problems \cite{jaderberg2016reinforcement} to improve learning. Here the auxiliary task is paired with an auxiliary input to indirectly control the output of a trained model.
    \item Posing increasingly difficult (progressive PCG) problems increase learning capacity \cite{justesen2018illuminating} much like curriculum learning. If the difficulty can be controlled by an auxiliary input we should be able to leverage this.
    \item Generally, we draw inspiration from the idea of adversarial learning in generative models \cite{goodfellow2014generative}, which leverage adversarial examples to train a more robust classifier. Here we generate adversarial environments to train a more robust solver.
\end{itemize}

As previously mentioned, our model consists of two co-existing adversarial agents where the Generator creates an environment (e.g. racing tracks, platforms, paths) which the Solver is tasked to solve/traverse. The Solver gives feedback to the Generator in the form of observations and rewards, and the Generator challenges the Solver by creating an adapted problem. This way the system is symbiotic, as without the Solver the Generator would not be able to create something that is "playable" (solved by a player), and the Solver without the Generator would not be able to generalize well over unseen environments. The use-case for this model is twofold. 1) Training a Generator to make the Solver fail makes the Solver more robust 2) The Generator can be used to create new environments which are shown to be traversable by the Solver. For inducing control and better diversity in the Generator's output, we introduce an auxiliary task (see section \ref{sec:generator} for more details). A schematic view of the architecture can be found in Fig. \ref{fig:ARLPCGModel}. We refer to this model as {\it adversarial reinforcement learning for procedural content generation} (ARLPCG). To test the ARLPCG model, we utilize two environment types; see section \ref{sec:environments} for more details. 

To summarize, this Solver-Generator model is designed to address a set of challenges:
\begin{itemize}
    \item The Generator should provide diverse training data allowing the Solvers to become more robust and to handle all/most environments authored either by the Generator or a human (e.g. game developer or player).
    \item The Generator can assist game designers to create environments that could be controlled and quantified by designed metrics (e.g. difficulty).
    \item The Solver can assist game designers to test environments in real-time in production.
\end{itemize}

In the following sections we describe the individual components of the model.

\subsection{The Generator}
\label{sec:generator}
Similarly to e.g. PCGRL \cite{khalifa2020pcgrl}, we focus on an iterative creation process as opposed to generating the whole environment at once. However, in contrast to PCGRL where the agent manipulates an existing randomly generated environment (where the amount of manipulation is a hyper-parameter), ARLPCG generates the whole environments from scratch in a iterative fashion where the environment starts empty and it is created segment per segment. One advantage with this approach is that the Generator does not create a new section/segment until the Solver reaches the latest section, thus always creating a solvable/reachable segment for the playing character. Furthermore, with the use of an auxiliary task we induce indirect control over the process. Reward for the Generator is partially deduced from the Solver, i.e. if the Solver makes progress the Generator also receives a reward in order to drive the creation of meaningful environments. As the Generator learns to make challenges that are difficult but not impossible for the Solver, the Generator can be used to also drive creation of new environments.

\subsubsection{Auxiliary input}
\label{sec:auxiliary_input}
When training the Generator, there is a balance to be struck between impossible and trivial environments. Furthermore, training an adversary RL based Generator against an RL based Solver agent will most likely lead to convergence towards the optimal utility function for both agents. This is undesirable in two ways: the solutions are often quite similar in nature leading to low generalization ability for the Solver, and the Generator model then allows for little control. Therefore, we introduce an auxiliary input to the Generator network. 

Our goal is to let an auxiliary input control the reward function in such a way that we can control the output indirectly. This way, the Solver is confronted with varying difficulty and behaviour (controlled by the auxiliary input to the Generator) which increases its generalization ability to solve previously unseen environments. Thus, the reward of the Generator is connected to auxiliary rewards based on the auxiliary input to the network. In this way, the behaviour of the Generator changes when the auxiliary input is changed. When training, the auxiliary input is randomly generated (range: $[-1, 1]$), see Fig. \ref{fig:ARLPCGModel}. Our hypothesis is that this will yield enough variability for the model to better generalize than scripted PCG environments. Herein, we focus on using the auxiliary input to control the degree of difficulty of the environment, but it could also be used to drive other metrics or behaviours, such as speed, actions required, etc. 

\subsubsection{Reward structure}
As previously mentioned, the Generator receives a reward from two sources. One is internal, meaning that it depends only on its own actions. The other one is the external reward which is tied to the performance of the Solver. In this paper we mainly connect the performance on progression and failure but it can be set differently depending on the desired behaviour of the Generator as will be shown in section \ref{sec:results}. 

For training a Generator to create a challenging and diverse environment, we design the reward functions to mainly drive two properties: progress and behaviour. On one end of the spectrum the Generator should help the Solver to reach the goal (progress), and on the other end it should actively try to make it behave "sub-optimally" (i.e. any deviation from the fastest-path to goal could be considered sub-optimal, but this is also where the behaviour is manifested). By connecting the reward function to the auxiliary input to the network, we introduce a method to control the behaviour of the Generator. The desired behaviour depends on the environment and can be controlled/designed by a game developer or similar.

Formally, this reward function with auxiliary scaling can be written as:

\begin{equation}
    r = \sum_{i=0}^n r_{int} \lambda_{A_i} \alpha_i + r_{ext} \sum_{i=0}^n \lambda_{A_i} \beta_i, 
\end{equation}
where $\lambda_{A_i}\in [-1, 1]$  is the auxiliary input fed in as input to the network, $r_{int}$/$r_{ext}$ are the internal/external rewards, and $\alpha_i, \beta_i$ are weighting factors. 

\subsection{The Solver}
This approach deploys an RL agent as the Solver but one could imagine scripted agents playing against the Generator. However, we argue that this would likely lead to the Generator finding exploits in the scripted agents rather than producing meaningful tasks (see section \ref{sec:results_generator} for further discussion). Furthermore, it would defeat the purpose of adaptiveness as a programmer would have to re-write the code when the environment changes significantly. Therefore, we argue, it is crucial that the Solver has the ability to learn and adapt. Hence, we choose to use an RL agent as the Solver in this model. The training hyper-parameters can be found in Table \ref{tab:SolverSetup}. Generally, the reward function for the Solver contains a progressive reward, plus a penalty for failing. The negative reward for failing is crucial as it indirectly forces the Generator to create environments that are not impossible to solve.

\section{Environments}
\label{sec:environments}

To validate our model we create two types of 3D environments, each representing two popular game genres: a third person platformer and a racing game. The Generator parameterizes the environment segment by segment by outputting the generating control parameters. The goal of the Solver is to traverse the generated environment as fast as possible without falling/driving off the track. The games were built in Unity and connected to the training environment using {\it Unity ML-Agents} API \cite{juliani2018unity}.

\begin{figure}[t!]
    \centering
    \begin{subfigure}[b]{0.24\textwidth}
        \centering
        \includegraphics[width=\textwidth]{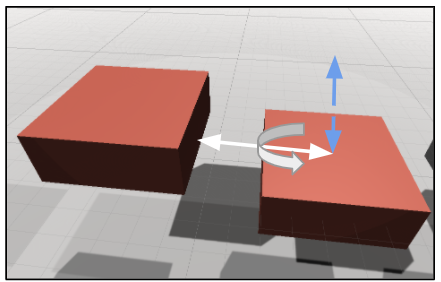}
        \caption{}    
        \label{fig:PlatformGeneratorExplained}
    \end{subfigure}
    \hfill
    \begin{subfigure}[b]{0.24\textwidth}  
        \centering 
        \includegraphics[width=\textwidth]{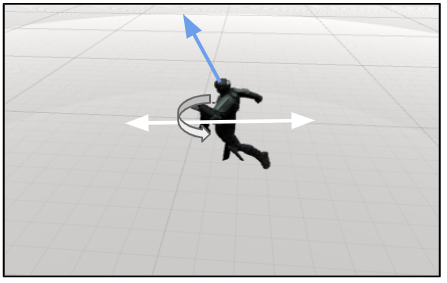}
        \caption{}    
        \label{fig:PlatformSolverExplained}
    \end{subfigure}
    
    \caption{Platform game. a) The Generator creates segments with actions: distance (white solid arrow), angle (gray arrow), and height (blue dotted arrow) relative to the previous block. The agent also controls the size of the block. b) The Solver: drives a character with actions: forward/backward (white solid arrow), turn (gray arrow), and jump (blue solid arrow).}
    \label{fig:PlatformGameExplained}
\end{figure}

\begin{figure}[t!]
    \centering
    \begin{subfigure}[b]{0.215\textwidth}
        \centering
        \includegraphics[width=\textwidth]{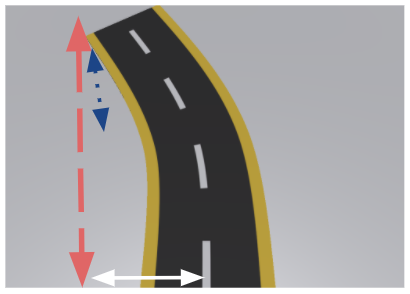}
        \caption{}    
        \label{fig:DriverGeneratorExplained}
    \end{subfigure}
    \hfill
    \begin{subfigure}[b]{0.235\textwidth}  
        \centering 
        \includegraphics[width=\textwidth]{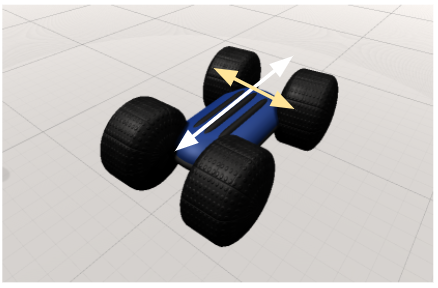}
        \caption{}    
        \label{fig:DriverSolverExplained}
    \end{subfigure}
    
    \caption{Driver game. a) The Generator creates segments with actions: length (red dashed arrow), turn (white solid arrow), and height (blue dotted arrow) as input. b) The Solver: drives a vehicle with actions: throttle (white solid arrow) and turn (yellow solid arrow).}
    \label{fig:DriverGameExplained}
\end{figure}

\subsection{Representation}
In our environments the action and observation space are local and relative to the previous object placed. We therefore apply a local representation scheme meaning that tiles and segments are placed in a sequence (corresponding to {\it Turtle} in \cite{khalifa2020pcgrl}), as opposed to where the objects can be placed anywhere in the environment (i.e. {\it Wide} in \cite{khalifa2020pcgrl}).

\subsection{Racing game}
To represent a racing game we created a vehicle driving scenario, here referred to as {\it Racing Game}. Here, the player/agent controls a vehicle and the goal is to drive along a generated track as fast as possible to maximize reward. The road is created in segments where the banking is dependent on the curvature of the road. The episode terminates if the driver leaves the track, reaches the end or if the maximum step is reached.

\subsubsection{Generator}
The Generator's actions control the length of the segments (in meters $[20, 30]$), curve (in degrees $[-30, 30]$), and height change (in meters $[-5, 5]$). The new section is requested by the Solver \SI{15}{\metre} before the road ends, thus generating a constant flow of new sections that create a race track. The Generator's observation consists of both a game state array (relative position to the goal, heading, angle relative to the goal, goal distance, and auxiliary input), and a ray cast array to give it information about the environment ahead. The ray casts can be used as a way of creating a track in an already filled environment where they can work as a collision detector. We introduce a ray cast array in order to allow the model to be deployed in an already existing environment with obstacles already in place. This way the ray casts can be used for the agent to learn to avoid obstacles while still creating a traversable environment. If the track is colliding with itself or other obstacles, the track ends and the episode terminates.

\subsubsection{Solver}
The controller inputs are throttle and turn. The Solver receives a positive reward for moving towards the goal, and completing the track. It gets a negative reward for failing (driving off-track or timing out without completing). The observation consists of both a game state array (relative position to the goal, heading relative to the goal, angular velocity, velocity, and rotation), and a ray cast array to give it information about the road ahead and the surroundings of the vehicle.

\subsection{Platform Game}
To represent the navigation aspect of Platform games, and even some first person shooter (FPS) games, we created one called {\it Platform Game}. Here the player controls a character which goal is to traverse along a generated track as fast as possible to maximize reward.

\subsubsection{Generator}
The Generator's actions control distance to next block (in meters $[5, 10]$), angle relative to the last two blocks (in degrees $[-180, 180]$), square block size (in meters $[4, 6]$), and height change (in meters $[-2, 2]$). The observation consists of a game state array (relative position to the goal, angle relative to the goal, goal distance, previous block position/size/rotation, and auxiliary input).

\subsubsection{Solver}
The controller inputs are forward/backward, left/right turn, and jump for the Solver. The Solver receives a positive reward for moving towards the goal, and completing the track. It gets a negative reward for failing (falling or timing out without completing).

\section{Training}
\label{sec:training}
The models are trained with PPO and a version of self-play (alternating Markov game) \cite{littman1996generalized}, i.e. the Generator and Solver are trained in an iterative fashion where the state of the other network is frozen (i.e. only running inference) when the training network is updating. To note here is that it is not a zero-sum game where the gain of one is the loss of the other proportionally, but rather a semi-collaborative game where there are elements of both competition and collaboration. In this way, there are lower risks for exploits being developed and more substantial/relevant challenges are posed. Due to the nature of the problem, the Solver (in both the Platform and Racing Game) works on a much higher frequency doing hundreds of actions per episode while the Generator's episode is completed in the order of dozens. Therefore, the Solver training switch occurs on a frequency of about a tenth of the Generator. 
Training parameters can be found in Tables \ref{tab:GeneratorSetup} and \ref{tab:SolverSetup}.
\begin{table}[!t]
\centering
\begin{tabular}{lll}
\hline
{\it }                  & Platform                  & Racing Game             \\ \hline
Observation             & State vector              & State vector + Ray cast \\
Actions                 & 4                         & 2                       \\
Learning rate           & 2e-4                      & 2e-4                    \\
Layers $\times$ units   & 2 $\times$  512           & 2 $\times$ 512          \\
$\gamma$                & 0.99                      & 0.99                    \\

\hline
\end{tabular}
\caption{The Generator's setup. }
\label{tab:GeneratorSetup}
\end{table}

\begin{table}[h!]
\centering
\begin{tabular}{lll}
\hline
{\it }                  & Platform                  & Racing Game               \\ \hline
Observation             & State vector + Ray cast   & State vector + Ray cast   \\
Actions                 & 4                         & 2                         \\
Learning rate           & 3e-4                      & 3e-4                      \\
Layers $\times$ units   & 2 $\times$  512           & 2 $\times$ 512            \\
$\gamma$                & 0.990                     & 0.998                     \\
\hline
\end{tabular}
\caption{The Solver's setup. In both environments, the Solver uses ray casts to navigate around obstacles. In the Racing Game, the ray casts "fan out" around the vehicle to keep track of the road, and in the Platform game a height map around the agent is used.}
\label{tab:SolverSetup}
\end{table}

\subsection{Reward function and auxiliary input in training}
In our Platform Game experiments the Generator receives a reward of $\lambda_{A_i} \times 10$ whenever the Solver fails. This way we can encourage different behaviour of the trained Generator depending on the auxiliary input ($[-1, 1]$). See Figures \ref{fig:PlatformTrackHighAux} and \ref{fig:PlatformTrackLowAux} for examples of the difference these auxiliary input can generate in behaviour.

In the Racing Game, the auxiliary reward $r_A$ is connected to failure in the same way. As an additional experiment, when $\lambda_{A_i} < 0$ a positive reward is added for each time step the vehicle is above a certain threshold above ground. As a consequence, $\lambda_{A_i} = -1$ will maximize the air-time of the vehicle by heavily undulating the track, while $\lambda_{A_i} = 1$ will reward the Generator when the Solver moves towards the goal. See Figures \ref{fig:DriverTrackHighAux} and \ref{fig:DriverTrackLowAux} for examples of the difference these auxiliary inputs can generate in behaviour. This shows that we can control other features that are not connected to failure/success. 
When training in both environments, we found that randomly selecting auxiliary input values from ${-1, -1, -0.5, 0.5, 1, 1}$ gave a distribution that led to stable results compared to using random sampled values in the range $[-1, 1]$. The latter approach still lead to convergence and good results but it took longer to train generally. We believe it helps to expose the agents to edge values (i.e. -1 and 1) thus increasing exposure to auxiliary values that are far from each other that is beneficially for the mapping of auxiliary input to reward. Interestingly, we found that a fully trained model could extrapolate between these values reasonable well, see section \ref{sec:results_generator} for more details.

In both environments when the auxiliary input is negative, the Generator receives a small negative reward per step. The idea behind this setup is that this forces the Generator to create an environment that the Solver either (depending on the auxiliary task) finishes, or fails, fast. Thus, if the auxiliary value is low, the Generator will design a difficult environment while if the auxiliary value is high it should make it easy to traverse. 
Independent of the auxiliary input, the Generator receives an incremental reward for the Solvers progress, i.e. when the Solver gets closer to the predefined goal. In the training, the goal positions are randomized to further promote diversity, but also to train the Generator to create a path to a predefined position set by a game designer. The training hyper-parameters can be found it Table \ref{tab:GeneratorSetup}.


%

\section{Results}
\label{sec:results}
We are reporting results on mainly two research questions: 1) Can the Generator be used to create environments with different difficulties/behaviours which can be controlled with an auxiliary input? 2) Are the agents trained with the ARLPCG model better at generalization than agents trained with other approaches? We use two baselines to compare and validate our model: One is where the RL agent is trained on a fixed set of environments, similar to the Atari game suite. The other is similar to \cite{khalifa2020pcgrl} where PCG environments are generated based on a set of rules and random variables.

\subsection{Generator}
\label{sec:results_generator}
\begin{figure}[ht]
    \centering
    \begin{subfigure}[b]{0.24\textwidth}
        \centering
        \includegraphics[width=\textwidth]{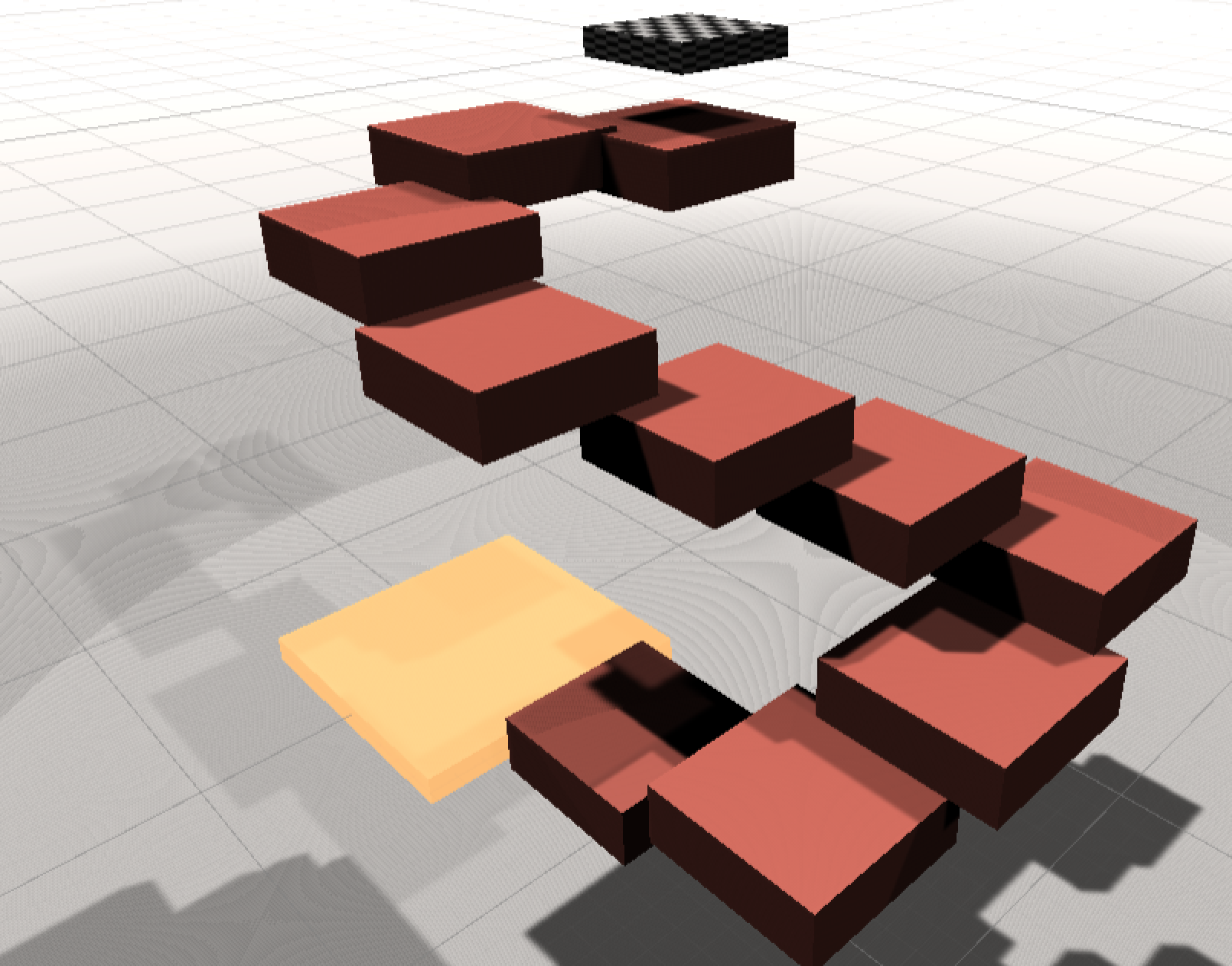}
        \caption{Platform Game.}    
        \label{fig:GeneratorStaircaseExample}
    \end{subfigure}
    \hfill
    \begin{subfigure}[b]{0.24\textwidth}  
        \centering 
        \includegraphics[width=\textwidth]{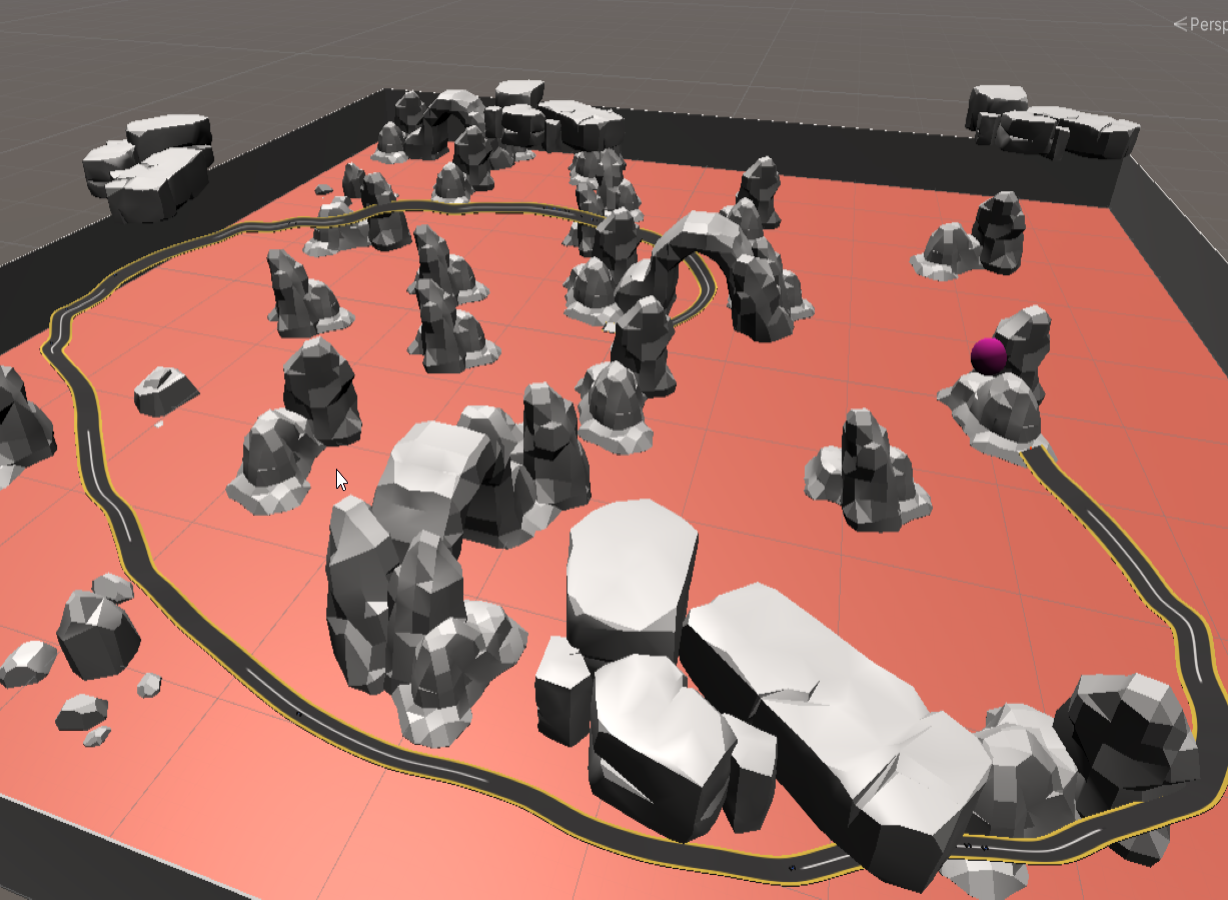}
        \caption{Racing Game.}    
        \label{fig:DriverGeneratorObstacles}
    \end{subfigure}
    
    \caption{Examples of generated environments. (a) Here the Generator had to create a path to a goal far above (\SI{20}{\metre}) the starting point. (b) Here the Generator had to create a path (between 4 waypoints) avoiding fixed obstacles in the environment.  }
    \label{fig:EnvironmentExamples}
\end{figure}

In both game environments, the Generator learned to create a path towards a randomly generated goal at the same time as it allowed the Solver to traverse it. In the Platform Game it was able to create complicated tracks that (first) led away from the goal in order to assure that the Solver would be able to jump up onto each platform. Much like a spiral staircase, it was able to create a vertical path for the agent if the goal was straight above the agent, see Figure \ref{fig:GeneratorStaircaseExample}. In the case of the Racing Game we also saw that it could be used in a non-empty environment where the ray casts made it possible to create a track around obstacles, see Figure \ref{fig:DriverGeneratorObstacles}. 

\begin{figure}[ht]
    \centering
    \begin{subfigure}[b]{0.23\textwidth}
        \centering
        \includegraphics[width=\textwidth]{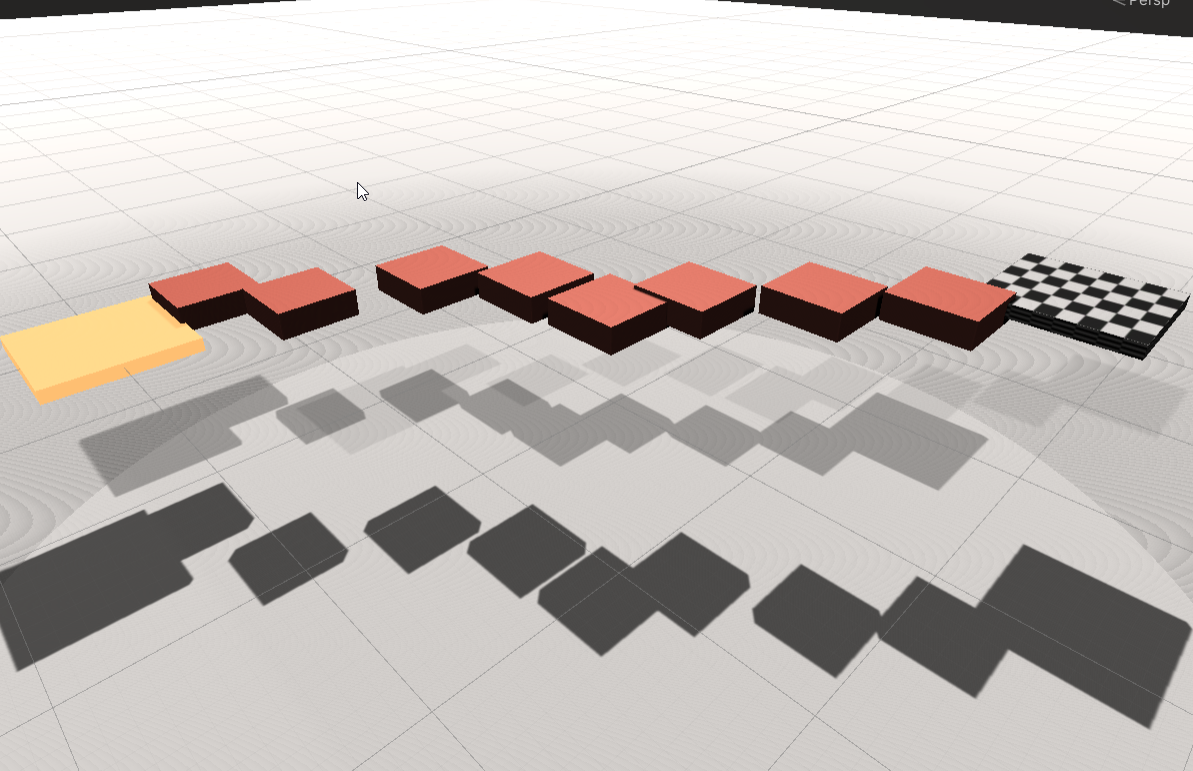}
        \caption{Auxiliary input of value 1.}    
        \label{fig:PlatformTrackHighAux}
    \end{subfigure}
    \hfill
    \begin{subfigure}[b]{0.24\textwidth}  
        \centering 
        \includegraphics[width=\textwidth]{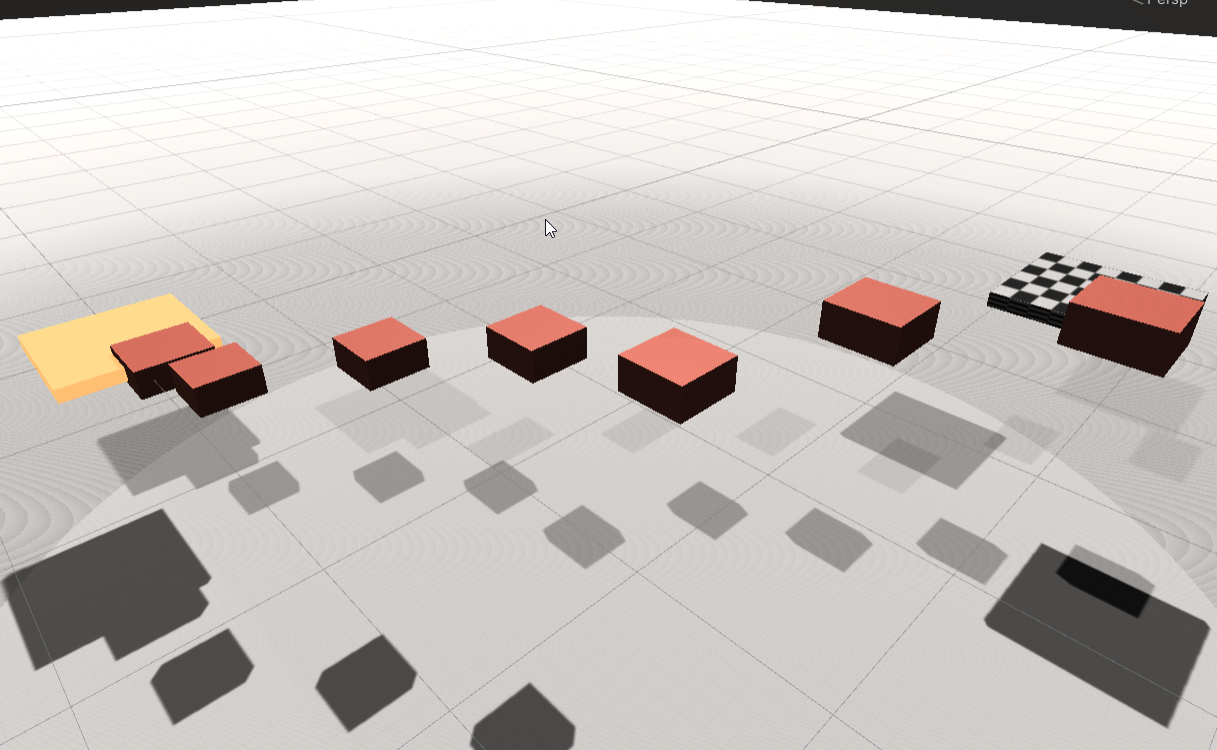}
        \caption{Auxiliary input of value -1.}    
        \label{fig:PlatformTrackLowAux}
    \end{subfigure}
    
    \caption{Examples of generated tracks with different auxiliary inputs to the Generator. With a high auxiliary input ($\lambda_{A_i}>0$), the Generator agent creates a track that is easy for the Solver to follow when trying to reach the waypoints (left figure). With a low auxiliary value ($\lambda_{A_i}<0$), it tries to make the Solver fail by creating a track that has long and difficult jumps (right figure).}
    \label{fig:PlatformExamples}
\end{figure}

To evaluate the value of having an adaptive agent, we simulated a scripted agent by deploying a trained RL agent and keeping its network fixed. As one would expect without the adapting Solver, the Generator with $\lambda_{A_i} = -1$ did converge towards an impossible track by creating jumps that the non-adaptive agent would continuously fall for.
However, replacing the fixed RL agent with a constantly training RL Solver it then learned to not try to jump over impossible gaps between blocks. The converged Generator instead produced tracks with long jumps and smaller platforms (for low auxiliary values), and short jumps with bigger platforms (for high auxiliary values). Thus, the Generator when having a low auxiliary input value converged towards hard but not impossible jumps that encourages the Solver to try the jump and eventually fail. See Figure \ref{fig:PlatformExamples} for examples. This is quantified for Platform Game in Table \ref{tab:results_generator_traversability} where it shows that the Generator, in order to maximize reward with the corresponding auxiliary input, created bigger jumps and smaller platforms. Similarly, for the Driver Game, Table \ref{tab:results_generator_driver} shows that lower auxiliary input gives sharper turns and more varying height difference between segments. Figure \ref{fig:PlatformTrackHighAux} and \ref{fig:PlatformTrackLowAux} shows the difference between a typical high and low auxiliary input. Here we can observe that without explicitly being told, the Generator creates tracks with bigger jumps (i.e. larger distance and smaller blocks) as it increases the chances of the Solver failure. Even for auxiliary values not used in training, the trend is that the tracks become increasingly harder to solve, i.e. it possible for it to extrapolate between values. 



\begin{figure}[ht]
    \centering
        \begin{subfigure}[b]{0.22\textwidth}
        \centering
        \includegraphics[width=\textwidth]{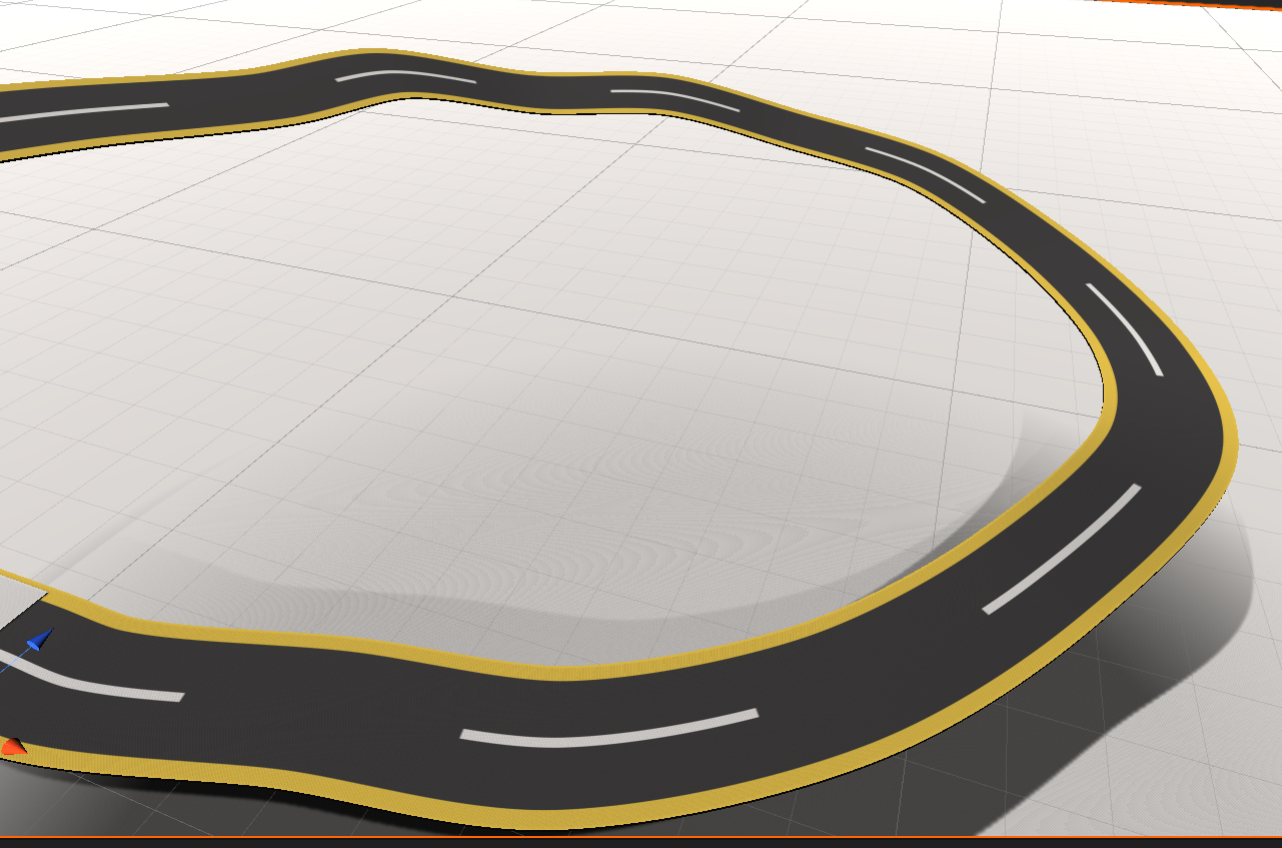}
        \caption{Auxiliary input of value 1.}    
        \label{fig:DriverTrackHighAux}
    \end{subfigure}
    \hfill
     \begin{subfigure}[b]{0.23\textwidth}  
        \centering 
        \includegraphics[width=\textwidth]{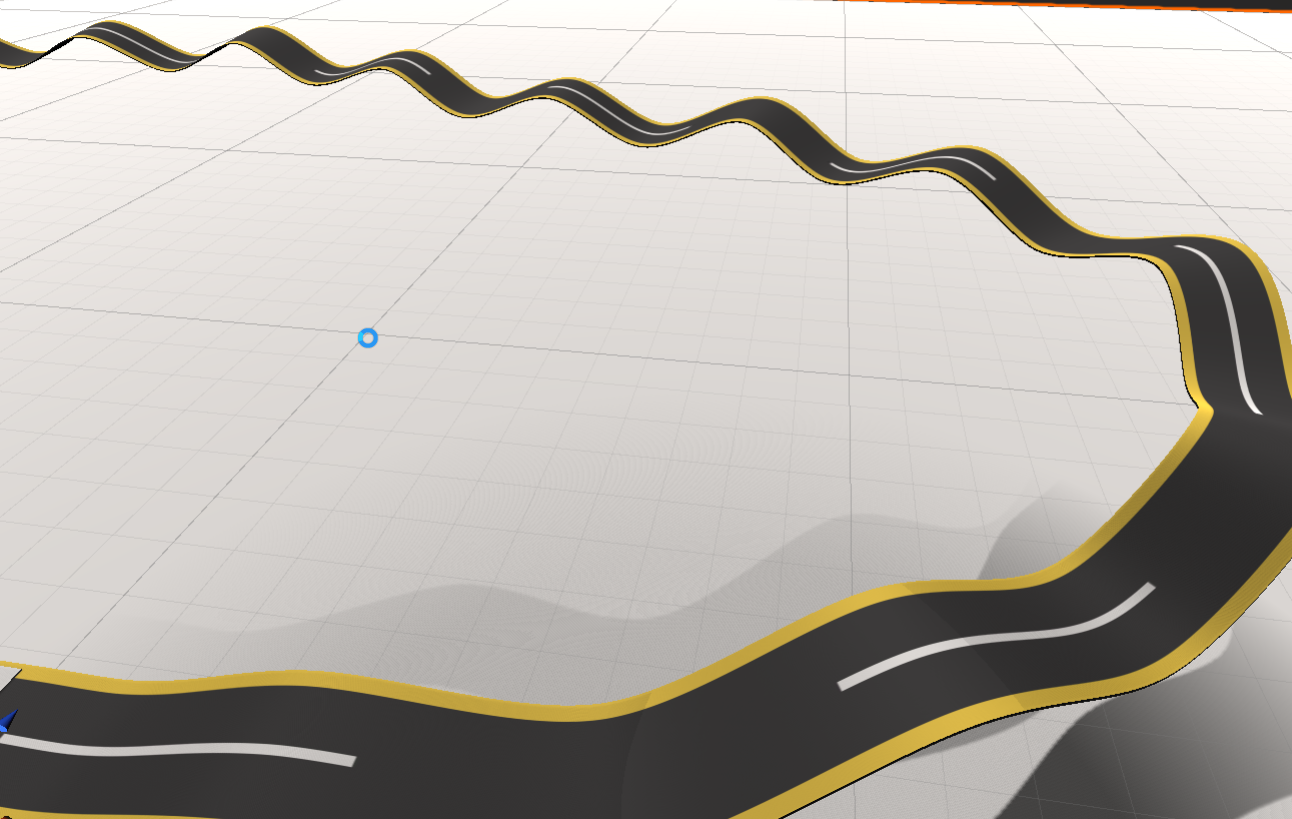}
        \caption{Auxiliary input of value -1.}    
        \label{fig:DriverTrackLowAux}
    \end{subfigure}
    
    \caption{Examples of generated tracks with different behaviour controlled via the auxiliary input. With a high auxiliary input ($\lambda_{A_i}>0$), the Generator agent creates a track where the Solver easily can get to the waypoints (left figure). With a low auxiliary value ($\lambda_{A_i}<0$), it is conditioned for reward for increases air-time causing the Generator to create a highly undulated track (right figure).}
    \label{fig:DriverExamples}
\end{figure}

\subsubsection{Generating environments with different style}
One benefit with our model is that the Generator can be used to generate human playable environments in different styles. By playing them, qualitatively there are noticeable differences between environments generated with auxiliary inputs ($\lambda_{A_i}$) of -1, 0, and 1 where the former value generates the harder ones.

Setting a high positive reward for the Generator when the Solver fails will result in generally more difficult tracks. However, the Generator will generate an environment that is "one-dimensional" in the sense that not everyone (both human and synthetic players) find the same difficulty equally challenging. Here, in the case of the Platform Game the distance for the jumps are longer (see Table \ref{tab:results_generator_traversability}). Similarly, in the Racing Game there is a tendency of the Generator to create a curvy road with negative camber (i.e. the track will be lower on the outer side of the curve) that causes the Solver to slide off if it enters the curve too fast. To investigate if we could control the way the difficulty was constructed we created a reward function which also rewarded air-time. The goal was to see if it was possible to both control difficulty while also give direction on {\it how} the difficulty should manifest. The results where promising and we could see a trend towards control over difficulty and style as well, see Table \ref{tab:results_generator_driver} and Figure \ref{fig:DriverExamples}.

To investigate the relation between the auxiliary input and the trained models output, we take the three different agents and validate them in a set of Generator generated tracks. For the ARLPCG Solver this can be seen as a function of the training where the Generator learn to adapt to the Solvers behaviour. Therefore, this is especially interesting for the agents which have been trained on a completely different type of track generation. As we see in Table \ref{tab:results_solvers_driver} and \ref{tab:results_solvers_platform}, all types of trained Solvers struggle more with environments generated by low auxiliary input (the supposedly harder ones) than the environments with high auxiliary input. With these two analysis approaches we can conclude that the Generator can be used to create different styles of tracks to a certain degree.

\begin{table}[!ht]
\centering
\begin{tabular}{llll}
\hline
$\lambda_{A_i}$ & {\it Success ratio} & {\it Avg. block distance} & {\it Avg. block size}\\ \hline
{1}             & 0.97     & \SI{6.73}{\metre} & \SI{5.95}{\metre} \\
{0.5}           & 0.92     & \SI{6.98}{\metre} & \SI{5.77}{\metre} \\
{0}             & 0.88     & \SI{7.15}{\metre} & \SI{5.42}{\metre} \\
{-0.5}          & 0.74     & \SI{7.11}{\metre} & \SI{4.52}{\metre} \\
{-1}            & 0.69     & \SI{7.46}{\metre} & \SI{4.01}{\metre} \\
\hline
\end{tabular}
\caption{Results on Generator creating environments in the Platform Game with with variable auxiliary input. Second column shows success ratio for Solver with different auxiliary input. Third and fourth columns show the average generated distances and block sizes.}
\label{tab:results_generator_traversability}
\end{table}

\begin{table}[!ht]
\centering
\begin{tabular}{lllll}
\hline
$\lambda_{A_i}$ & {\it Success ratio} & {\it Avg. Angle} & {\it Avg. Length}& {\it Avg. Height}\\ \hline
{1}             & 0.99     & \SI{14.4}{\metre} & \SI{28.4}{\metre} & \SI{1.5}{\metre} \\
{0.5}           & 0.89     & \SI{14.7}{\metre} & \SI{27.1}{\metre} & \SI{2.0}{\metre}\\
{0}             & 0.77     & \SI{15.1}{\metre} & \SI{25.1}{\metre} & \SI{2.5}{\metre}\\
{-0.5}          & 0.64     & \SI{15.3}{\metre} & \SI{23.0}{\metre} & \SI{3.1}{\metre}\\
{-1}            & 0.61     & \SI{16.8}{\metre} & \SI{21.8}{\metre} & \SI{3.9}{\metre}\\
\hline
\end{tabular}
\caption{Results on Generator creating environments in the Driver Game with with variable auxiliary input. Second column shows success ratio for Solver with different auxiliary input. Third, fourth and fifth columns show average change in angle, length and height, respectively.}
\label{tab:results_generator_driver}
\end{table}

\subsection{Solver}
\label{sec:results_solver}

To test the generalization ability, we did experiments by training the same RL agent (i.e. same hyper-parameters, observations, actions, rewards, etc.) on differently generated environment. We used three approaches: 1) Training on a fixed map (referred to as {\it Fixed Track}), 2) Training on a rule based PCG environment (referred to as {\it Rule PCG}), i.e. setting rules for the Generator and then generating environments based on random parameters and 3) ARLPCG where Solvers were trained on Generators with different auxiliary inputs. Results can be seen in Table \ref{tab:results_solver}. 

A conclusion we can draw from this is that, generally, ARLPCG solves previously unseen environments better than the other approaches, especially comparing to training on a fixed set of environments. In Table \ref{tab:results_solver} we see that in the Platform game the solve ratio is 0.457 compared to 0.233 (Rule PCG) and 0.081 (Fixed Track), and in the Racing game the solve ratio is 0.331 compared to 0.171 (Rule PCG) and 0.219 (Fixed Track). Interestingly, when it comes to the {\it Platform} game the steps required (on average) for {\it Fixed Track} to succeed is much lower than ARLPCG. The reason for this is that when it did succeed to get to the goal, it did quickly with very little hesitation. Similarly, when it failed it also did so with very "high confidence" by rushing straight out into thin air. It shows that it has, to a much larger degree, memorized certain action sequences. Usually the failures of the two PCG approaches were more subtle in the sense that failure mostly occurred on missed (but reasonable) jumps between blocks. 

We did an ablation study by omitting auxiliary inputs and rewards. The setup was similar but the auxiliary input was fixed to a value instead ($\lambda_{A_i} = 1$) of randomly sampled effectively causing the Generator to create similar style of environment each time. We found that the generalization ability was greatly reduced when training on these kinds of environments, see Table \ref{tab:results_solver} and row {\it Fixed Aux.}.

\begin{table}[ht]
\centering
\begin{tabular}{lll}
\hline
{\it }                & {\it Platform Game}                & {\it Racing Game}  \\ \hline
\textbf{Fixed Track}  & 0.081$\pm$0.205, 0.20 (54 steps)  & 0.219$\pm$0.280, 0.7 (431 m) \\
\textbf{Rule PCG}     & 0.233$\pm$0.200, 0.95 (451 steps) & 0.174$\pm$0.160, 0.9 (355 m) \\
\textbf{Fixed Aux.}   & 0.173$\pm$0.195, 0.90 (442 steps) & 0.163$\pm$0.138, 0.8 (382 m) \\
\textbf{ARLPCG}       & 0.457$\pm$0.211, 1.00 (467 steps) & 0.331$\pm$0.225, 1.0 (507 m) \\
\hline
\end{tabular}
\caption{Comparison of performance on a set of previously unseen validation tracks (1000 $\times$ 20 runs). {\it Platform Game} values indicate success rate, fraction of tracks completed by at least one agent, and average steps taken to reach the goal. {\it Racing Game} values indicate success rate, fraction of track completed, and average distance completed. {\it Fixed Track} is trained on a fixed set of tracks, {\it Rule PCG} is trained on rule based PCG generated from a set of rules randomized to create different track every time, and {\it Fixed Aux.} is trained on a Generator with a constant auxiliary input. }
\label{tab:results_solver}
\end{table}

\begin{table}[h]
\selectfont
\centering
\begin{tabular}{llll}
\hline
\textbf{Aux. value} & \textbf{Fixed Track}      & \textbf{Rule PCG}       & \textbf{ARLPCG} \\ 
\hline


1 (Easy)     & 0.743$\pm$0.369 & 0.788$\pm$0.301 & 0.997$\pm$0.009 \\
0.5          & 0.507$\pm$0.402 & 0.632$\pm$0.422 & 0.890$\pm$0.249  \\
0 (Moderate) & 0.280$\pm$0.349 & 0.622$\pm$0.425 & 0.772$\pm$0.340 \\
-0.5 (Hard)  & 0.281$\pm$0.336 & 0.460$\pm$0.441 & 0.643$\pm$0.392 \\
-1 (Hardest) & 0.206$\pm$0.322 & 0.425$\pm$0.444 & 0.613$\pm$0.430 \\

\hline
\end{tabular}
\caption{\textbf{Racing Game:} Solvers (trained on Fixed Track, Rule PCG, and ARLPCG environments respectively) average success ratio and speed on Generator generated tracks. Averaged over 100000 trials (250 trials on 50 tracks each \SI{500}{\metre} long). The Generator's auxiliary input is varied between [-1,1] to moderate the difficulty of the track. The success ratio and average speed reflects the difficulty even for the agents that were not trained using the Generator.}
\label{tab:results_solvers_driver}
\end{table}

\begin{table}[ht]
\selectfont
\centering
\begin{tabular}{llll}
\hline
\textbf{Aux. value} & \textbf{Fixed Track}  & \textbf{Rule PCG} & \textbf{ARLPCG}  \\ 
\hline
1 (Easy)            & 0.0$\pm$0.0           & 0.776$\pm$0.154   & 0.824$\pm$0.185  \\
0.5                 & 0.0$\pm$0.0           & 0.692$\pm$0.350   & 0.741$\pm$0.285  \\
0 (Moderate)        & 0.0$\pm$0.0           & 0.386$\pm$0.235   & 0.728$\pm$0.284  \\
-0.5 (Hard)         & 0.0$\pm$0.0           & 0.148$\pm$0.165   & 0.360$\pm$0.276  \\
-1 (Hardest)        & 0.0$\pm$0.0           & 0.053$\pm$0.037   & 0.183$\pm$0.139  \\
\hline
\end{tabular}
\caption{\textbf{Platform Game:} Solvers (trained on Fixed Track, Rule PCG, and ARLPCG environments respectively) average success ratio  on Generator generated tracks. Averaged over 50000 trials (50 tracks á 1000 trials). The Generator's auxiliary input is varied between [-1,1] to moderate the difficulty of the track.}
\label{tab:results_solvers_platform}
\end{table}

\section{Conclusion and Future Work}
\label{sec:conclusion}

The model presented shows that an RL based content generator can be trained to generate traversable environments that improves generalization of a solving RL agent. Without using an adaptive Solver, the Generator converges towards generating impossible environments as a non-adaptive solving agent is easily exploited. However, with an adaptive Solver, the Generator have to adapt its behaviour to create environments that are difficulty (to maximize reward when the Solver fails) but not too difficult to not discourage the Solver from even trying to traverse the environment. Therefore, pairing a Generator with an RL agent Solver is crucial for functionality. We expect other adaptive agent methods (other than RL) to give similar results, but such an analysis is out of scope for this paper.
We also note that without the auxiliary inputs, we get a lower generalization ability, albeit similar to rule based PCG which still works well. Using a multi-dimensional auxiliary input function could potentially increase the diversity of the generated environments. Another advantage with this approach is that the Generator can generate different environments (controlled via an auxiliary input) for other use cases, such as real-time map creation.

Confirming the findings in \cite{khalifa2020pcgrl}, the agents trained in a PCG environment perform better than agents trained on a fixed set of maps. We also saw that certain environments benefit more from training on diverse environments like the agents in the Platform Game (compare Table \ref{tab:results_solvers_driver} and \ref{tab:results_solvers_platform}). We reason that the Driver environment is much less complex as driving along a spline can be generalized down to a few rules, while jumping on platforms the agents may potentially face a larger set of novel situations.  Furthermore, we show that we can improve generalization compared to this approach by using an adversarial RL architecture which constantly challenges the RL agent with adapted PCG environments. 

We argue that the model's two parts are symbiotic in the sense that they are both dependent on the other to develop and evolve their behaviours. In our case, using a high auxiliary value (i.e. giving reward to the Generator for the progress of the Solver) they work together maximizing each other's reward, while with a low auxiliary value the Generator pushes the Solver to adapt and thus becoming more robust to change.

Here we tried two kinds of behaviours: One that only focused on difficulty (i.e. Platform game where reward was only connected to failure/success) and one that focused on map design (i.e. Racing Game where reward was connected to the behaviour of the Solver). We do not foresee any conceptual problems exploring other kinds of behaviours if a meaningful auxiliary input and corresponding reward function can be designed.

In this paper, we chose the auxiliary rewards to showcase certain behaviours of the Generator showing that this approach is viable. We believe that other types of auxiliary functions (i.e. multi-dimensional and/or multi-behavioural) could be used to increase the generalization of the Solver and behaviour of the Generator. Furthermore, another possible improvement of this approach could be instead of using one Solver per Generator, using a population of Solvers could further improve the diversity of the Generator's output as it would be harder for it to exploit a behaviour.

\bibliography{main}
\bibliographystyle{IEEEtran}

\end{document}